# Dragonfly Algorithm and its Applications in Applied Science - Survey

Chnoor M. Rahman[1,2], and Tarik A. Rashid[3]

[1] Technical College of Informatics, Sulaimany Polytechnic University, Sulaimany, Iraq

[2] Applied Computer Department, College of Health and Applied Sciences, Charmo University, Sulaimany, Iraq.

[3] School of Science and Engineering, Science and Engineering Department, University of Kurdistan Hewler, Erbil, Iraq.

**Abstract** One of the most recently developed heuristic optimization algorithms is dragonfly by Mirjalili. Dragonfly algorithm has shown its ability to optimizing different real world problems. It has three variants. In this work, an overview of the algorithm and its variants is presented. Moreover, the hybridization versions of the algorithm are discussed. Furthermore, the results of the applications that utilized dragonfly algorithm in applied science are offered in the following area: Machine Learning, Image Processing, Wireless, and Networking. It is then compared with some other metaheuristic algorithms. In addition, the algorithm is tested on the CEC-C06 2019 benchmark functions. The results prove that the algorithm has great exploration ability and its convergence rate is better than other algorithms in the literature, such as PSO and GA. In general, in this survey the strong and weak points of the algorithm are discussed. Furthermore, some future works that will help in improving the algorithm's weak points are recommended. This study is conducted with the hope of offering beneficial information about dragonfly algorithm to the researchers who want to study the algorithm.

**Keywords** Metaheuristic Algorithms, Optimization Algorithms, Swarm Intelligence, Single objective Optimization, Multi-objective Optimization, DA, Dragonfly Algorithm

## I. Introduction

Computational Intelligence (CI) is one of the newest areas of research. CI is a set of methodologies inspired by nature. CI based techniques are useful to solve complex real-world problems when the traditional methods are ineffective. Fuzzy logic, artificial neural network, and evolutionary computation are part of CI.

Evolutionary computation mainly concerns optimization problems including combinatorial, mixed or continuous problems. Evolutionary strategies and genetic algorithms are examples of evolutionary computation. However, this field has extended its scope to cover other areas.

Swarm intelligence (SI), for example, is part of the evolutionary computation. Nevertheless, the efficiency of SI based algorithms has attracted many researchers in various areas, which makes SI become a separate field [1]. Swarm-based algorithms produce low cost, fast, and robust solutions to complex real world problems [2]. In SI based techniques a number of agents form a population. Agents in a population interact each other and their environment. Nature (particularly, biological systems) is a great inspiration for these algorithms [3]. In SI techniques, agents practice simple rules. General control structures do not exist to show how individuals should behave. Interaction between agents causes the disclosure of global intelligent behaviour, which is not known to the agents [4]. Recently many SI based algorithms have been proposed. Most of them are mimicking the swarm and animal behaviours in nature. The most popular SI algorithms include particle swarm optimization (PSO) proposed by Kennedy and Eberhart [5]. PSO can be counted as a significant improvement in the field. It mimics the behaviors of the school of birds or fish. A particle represents a single solution that has a position in the search space. Furthermore, at the beginning of the 1990s, Marco Dorigo completed his Ph.D. thesis on optimization and nature-inspired algorithms. In his thesis, he examined a novel idea known as an ant colony optimization algorithm (ACO) [6]. Chu and Tsai developed Cat Swarm Optimization algorithm (CSO) based on the behaviour of cats [7]. Grey wolf optimizer (GWO) introduced by Mirjalili et al. [8]. GWO mimics the hunting behaviour of wolfs. Later, dragonfly optimization algorithm (DA) proposed by the same author [9]. DA was mainly inspired by the hunting and migration behaviours of dragonfly. Another example is the differential evolution algorithm (DE) [10]. DE is a direct search population based technique, stimulated by the evolution of living species. In [11] artificial bee colony (ABC) was proposed. ABC mimics the behaviors of honeybees. The results of this algorithm proved that it has well-balanced exploitation and exploration ability. Fitness dependent optimizer (FDO) proposed in [12]. FDO inspired by bee swarming reproductive process. However, it does not have any algorithmic connetion with the artificial bee colony algorithm or the honey bee colony algorithm. Instead, it is based on the PSO. Donkey and Smuggler Optimization Algorithm (DSO) was proposed in [13]. DSO mimicks the searching behaviors of donkeys. Searching and selecting routes by donkeys were utilized as an inspiration of the algorithm. Firefly algorithm (FA) [14], is another metaheuristic algorithm. It simulates the flashing behaviour of fireflies and the fact of bioluminescent communication. FA is counted as one of the most powerful techniques for solving the constrained optimization problems and NP-hard problems.

The importance of the metaheuristic algorithms and the reason that they have been used in many applications has encouraged the researchers to publish survey papers on the algorithms. For example a systematic and meta-analysis survey of whale optimization algorithm [15]. In [16] a survey on the new generation of metaheuristic algorithms was presented. Another survey on nature inspired meta-heuristic algorithms with its domain specifications was proposed in [17]. In [18] the recent development on the modifications of the cuckoo search (CS) algorithm was proposed. The CS is originaly developed by Yand and Dep [19]. It mimicks the brood parasitic behaviour of cuckoo species and utilized the Levy flight action of some fruit flies and birds.

One of the most recently swarm based algorithms is the dragonfly algorithm. It has been successfully utilized in many different applications. The DA is found to produce competitive and efficient results in almost all the applications that utilized it. After publishing the algorithm in 2016 until the end of working on this survey, it has been utilized to optimize a lot of problems in different areas. Thus, this paper is centered to review dragonfly algorithm as one of the most recent algorithms in the area.

This work first presents an overview of the DA. The variants of the algorithm are then described. Furthermore, the hybridization versions of the algorithm with other algorithms are addressed. Additionally, applications in the applied science fields are discussed. Moreover, a comparison between the DA and some other metaheuristics is made. The advantages and disadvantages of DA are then discussed. The DA algorithm is also tested on the CEC-C06 2019 benchmark functions. Furthermore, the PSO, DE and FA are tested on the traditional benchmark functions and the results are shown and compared with the results of the DA. In addition, a discussion and some problems of DA are presented along with providing solutions and future works to make the algorithm work better. Finally, a conclusion is given.

## II. Overview of DA

DA is mimicking the swarming behaviours of a dragonfly. The reason for their swarming is either migration or hunting (dynamic swarm or static swarm respectively). In static swarm, small groups of dragonflies move over a small area to hunt other insects. Behaviours of this type of swarming include local movements and abrupt changes. In dynamic swarming, however, a massive number of dragonflies create a single group and move towards one direction for a long distance [20]. The aforementioned swarming behaviours are counted as the main inspiration of DA. Static and dynamic swarming behaviours are respectively in line with the exploration and exploitation phases of metaheuristic optimization algorithm. Fig. 1 shows the behaviours of dragonflies in static and dynamic swarming. To direct artificial dragonflies to various paths five weights were used, which are separation weight ($s$), alignment weight ($a$), cohesion weight ($c$), food factor ($f$), enemy factor ($e$), and the inertia weight ($w$). To explore the search space high alignment and low cohesion weights are used, however, to exploit the search space low alignment and high cohesion weights can be used. Furthermore, to transfer between exploration and exploitation the radii of neighbourhood enlarged proportionally to the number of iterations were used. Tuning the swarming weights ($s, a, c, f, e,$ and $w$) adaptively during the optimization process is another way to balance exploration and exploitation. Mathematically each of the aforementioned weight factors shown in Equations (1) to (5).

The separation can be calculated as mentioned by Reynolds [21]:

$$S_i = -\sum_{j=1}^{N} X - X_j \qquad (1)$$

In equation (1), $X$ indicates the position for the current individual. $X_j$ is the position for the $j^{th}$ neighbouring dragonfly. And $N$ is the number of individual neighbours of the dragonfly swarm. And $S$ indicates the separation motion for the $i^{th}$ individual.

Equation (2) was used for calculating alignment [9]:

$$A_i = \frac{\sum_{j=1}^{N} V_j}{N} \qquad (2)$$

Where $A_i$ is the alignment motion for $i^{th}$ individual. $V$ is for the velocity of the $j^{th}$ neighbouring dragonfly.

Cohesion was expressed as follows:

$$C_i = \frac{\sum_{j=1}^{N} X_j}{N} - X \qquad (3)$$

Where $C_i$ is the cohesion for $i^{th}$ individual. $N$ is the neighbourhood size, $X_j$ is the position of the $j^{th}$ neighbouring dragonfly, and $X$ is the current dragonfly individual.

Attraction motion towards food is computed as follows:

$$F_i = X^+ - X \qquad (4)$$

Where $F_i$ is the attraction of food for $i^{th}$ dragonfly, $X^+$ is the position of the source of food, and $X$ is the position of the current dragonfly individual. Here, the food is the dragonfly that has the best objective function so far.

Distraction outwards predators are calculated as follows:

$$E_i = X^- + X \qquad (5)$$

Where $E_i$, is the enemy's distraction motion for the $i^{th}$ individual, $X^-$ is the enemy's position, and $X$ is the position of the current dragonfly individual.

For position updating in the search space, artificial dragonflies use two vectors: step vector $\Delta X$ and position vector $X$. the step vector is analogy to velocity vector in the PSO algorithm [5]. The position updating is also based mainly on the PSO algorithm framework. The step vector is defined in [9] as follows:

$$\Delta X_{t+1} = (sS_i + aA_i + cC_i + fF_i + eE_i) + w\Delta X_t \quad (6)$$

Where $S_i$ represents the separation for the $i^{th}$ dragonfly; $A_i$ is the alignment for $i^{th}$ dragonfly; $C_i$ represents the cohesion for $i^{th}$ dragonfly; $F_i$ represents the food source for the $i^{th}$ individual; $E_i$ represents the position of enemy for $i^{th}$ dragonfly; $w$ is the inertia weight; $t$ indicates the iteration counter.

When the step vector calculation is finished, the calculation for the position vectors start as follows:

$$X_{t+1} = X_t + \Delta X_{t+1} \quad (7)$$

Where $t$ indicates the current iteration.

In order to raise the probability of exploring the whole decision space by an optimization algorithm, a random move needs to be added to the searching technique. When no neighbouring solutions are there, to increase randomness, stochastic behaviour, and exploration of artificial dragonfly individuals, dragonflies are required to use a random walk (Lévy flight) to fly throughout the search space.

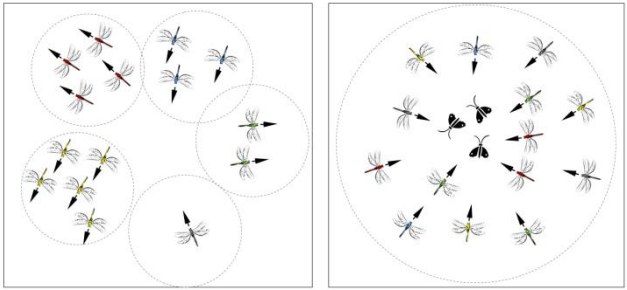

Figure 1: Dynamic dragonfly swarming (on the left hand) versus static swarming (on the right hand) [9]

## III. Convergences and Divergence Of DA

For the transition from intensification to diversification, dragonflies should adaptively change their weights. This guarantees the convergence of dragonfly individuals during the optimization process. As the optimization process progress, to adjust flying path the neighbourhood area is expanded, hence at the final stage of optimization, the swarm becomes one group to converge to a global optimum. The best and the worst solutions found so far become the food source and enemy respectively. This makes convergence and divergence towards the promising area and outwards non-promising area of the search space respectively.

## IV. Variants Of DA

DA has three variants:

### A. DA For Single Objective Problems

In DA, at the beginning of the process of the optimization process randomly a set of solutions is created. Initially the step and position vectors of artificial dragonflies are assigned to stochastic values between lower and upper bounds. The position and step vectors for each dragonfly should be updated in each iteration using Equations (7) or (8), and (6). To update step vector and position vector of dragonflies their neighborhood is chosen by Euclidean distance calculation. The position updating is continued until meeting the end criterion. Visual 1 shows the pseudo code for DA for single objective problems.

The single DA is the most popular variant among the other variants of DA.

---

*Initialize the dragonflies population Xi (i = 1, 2, ..., n)*

*Initialize step vectors Xi (i = 1, 2, ..., n)*

**while** *the end condition is not satisfied*

   *Calculate the objective values of all dragonflies*

   *Update the food source and enemy*

   *Update w, s, a, c, f, and e*

   *Calculate S, A, C, F, and E using Eqs. (1) To (5)*

   *Update neighbouring radius*

   **if** *a dragonfly has at least one neighbouring dragonfly*

     *Update velocity vector using Eq. (6)*

     *Update position vector using Eq. (7)*

   **else**

     *Update position vector using Lévy flight*

   **end if**

   *Check and correct the new positions based on the boundaries of variables*

**end while**

---

VISUAL 1. PSEUDO-CODE FOR DA [9]

### B. DA For Binary Problems

In binary search space, the position vector can take 0 or 1. Hence, the position of search agents cannot be updated by adding step vector to position vector. Using transfer function is the easiest way to convert continuous SI technique to binary algorithm [22]. Transfer function takes velocity (step) values as input and returns a number between 0 and 1 as output, which indicates the probability of changing the individual's position. Similar to continuous optimization, the function simulates sudden changes in particles with large velocity. To use the DA for binary problems (BDA) Equation (8) was used [22].

$$T(\Delta X) = \left| \frac{\Delta X}{\sqrt{\Delta X^2 + 1}} \right| \quad (8)$$

Equation (8) was used to calculate the changing probability of the position of all artificial dragonflies. Equation (9) for updating position was then created to update the search agent's position in binary search spaces.

$$X_{t+1} = \begin{cases} \neg X_t & r < T(\Delta X_{t+1}) \\ X_t & r \geq T(\Delta X_{t+1}) \end{cases} \quad (9)$$

Where $r$ is a number in [0, 1].

In BDA it was assumed that all of the artificial dragonflies are in one swarm. Therefore, BDA simulates exploration and exploitation by adaptively tuning the swarming factors ($s$, $a$, $c$, $f$, and $e$) and the inertia weight ($w$). Visual 2 presents the pseudo code for BDA.

> *Initialize the dragonflies population Xi (i = 1, 2, ..., n)*
>
> *Initialize step vectors Xi (i = 1, 2, ..., n)*
>
> **while** *the end condition is not satisfied*
>
>   *Calculate the objective values of all*
>
>   *dragonflies Update the food source and*
>
>   *enemy Update w, s, a, c, f, and e*
>
>   *Calculate S, A, C, F, and E using Eqs. (1) to (5)*
>
>   *Update step vectors using Eq. (6)*
>
>   *Calculate the probabilities using Eq (8) Update*
>
>   *position vectors using Eq. (9)*
>
> **end while**

VISUAL 2: PSEUDO-CODE FOR BDA [9]

## C. The DA For Multi-objective Problems

Multi-objective problems have multiple objectives. The result for multi-objective problems is a set called Pareto optimal set. The set contains the best trade-offs between the objectives [23].

In order to use the DA to deal with multi-objective problems (MODA), an archive was first provided to save and retrieve the best Pareto optimal solutions during the process of optimization. To update the position, the food source is selected from the archive, and the rest of the process of position updating is identical to that of DA.

Similar to the multi-objective particle swarm optimization (MOPSO) algorithm [24], to observe the well-spread Pareto optimal front, the food source is chosen from the least populated region of the produced Pareto optimal front. In MODA this was done through finding the worst and the best objectives of the current Pareto optimal solutions. Furthermore, a hypersphere to cover all the solutions was defined, and then in each iteration, the hyper-spheres are divided into equal sub-hyper-spheres. When the segments are created a roulette-wheel mechanism with the following probability for every segment was used for the selection process [25].

$$P_i = \frac{c}{N_i} \quad (10)$$

Where $c$ is a constant number and greater than one. $N_i$ is the number of Pareto optimal solutions obtained in the $i^{th}$ segment. Equation (10) gives the MODA a higher probability to select the food source from the less populated segments.

On the other hand, to select predators from the archive, the worst (most populated) hyper-sphere was chosen, so that the artificial dragonflies are prevented from searching around non-promising areas. For the selection process the roulette-wheel mechanism with the following probability was used:

$$P_i = \frac{N_i}{c} \quad (11)$$

Where $c$ is a constant number and greater than one. $N_i$ is the number of Pareto optimal solutions obtained in the $i^{th}$ segment.

In Equation (11), using the roulette-wheel mechanism; the most crowded hyper-spheres have a higher probability of being selected as enemies. To prevent the archive from becoming full, if at least one of the archive residences dominates the solution, then it should not be allowed to enter the archive. However, the Pareto optimal solutions dominated by the solution should be removed from the archive, and the solution should be added to the archive. If the archive is full, then one or more solutions may be removed from the most populated segments [25]. In addition to the parameters of DA, MODA has two new parameters; one for defining the maximum number of hyperspheres and another parameter to specify the archive size. The pseudo code for MODA is presented in Visual 3.

## V. Hybridization versions of DA

In the metaheuristic context, hybridization refers to merge the powerful characteristics of two or more algorithms in order to provide a new powerful algorithm based on the features of the merged ones [26]. In the following subsections, the hybridization versions of DA are discussed.

```
Initialize the dragonflies population Xi (i = 1, 2, ..., n)
Initialize step vectors      Xi (i = 1, 2, ..., n)
Define the maximum number of hyper spheres (segments) Define the
archive size
while the end condition is not satisfied
   Calculate the objective values of all dragonflies
   Find the non-dominated solutions
   Update the archive with respect to the obtained non-dominated
   solutions
   If the archive is full
      Run the archive maintenance mechanism to omit
      one of the current archive members Add the new
      solution to the archive
   end if
   If any of the new added solutions to the archive is
      located outside the hyperspheres
       Update and re-position all of the hyper spheres to cover the new
       solution(s)
   end if
   Select a food source from archive: =
   SelectFood(archive)
   Select an enemy from archive: = SelectEnemy(archive)
   Update step vectors using Eq. (8)
   Update position vectors using Eq. (9)
   Check and correct the new positions based on the
   boundaries of variables
end while
```

VISUAL 3: PSEUDO-CODE FOR MODA [9]

Reference [26] combined some features of PSO with DA and produced a new algorithm called memory based hybrid dragonfly algorithm (MHDA). In MHDA two more features are added to the DA to refine its performance, they are: (1) internal memory is added to observe the possible solution. This internal memory has a great role in converging to global optima. When the internal memory is added, each dragonfly individual will be able to keep track of its correlates in the problem space, which are related to the value of fitness. In the PSO algorithm, this is named as pbest. The best fitness value in each iteration is compared to the search agent's fitness value of the current population. As a result, the DA-pbest is created from the better-saved solutions. The dragonfly individuals are also able to keep track of the best value founded so far by any dragonfly in the neighbourhood. This is similar to the concept of gbest in the PSO. Here DA-gbest is used to store the best value. The capability of exploitation in DA is enhanced by these two novel concepts; pbest and gbest. The internal memory feature gives a greater performance comparing to the conventional algorithm and gives the power to escape from local optima [27]

(2) Iterative level hybridization with PSO, which runs on the saved solution's set. To enhance the performance of optimization in the iteration level hybridization approach two algorithms are executed iteratively in sequence [28]. Here to extend the search space and converge to a more promising area, DA with internal memory is used, and then the previously limited area is exploited using PSO to find better solutions.

Thus to reach global optimal solutions in the MHDA, the DA's exploration features in the initial stage and PSO's exploitation features in the final stage were combined. Visual 4 shows the pseudo-code of MHDA. MHDA's superior performance on unimodal functions showed speed converge and an accurate diversification of the algorithm. The results of the algorithm showed the competitive performance of the algorithm and that it can be used to optimize hard problems.

Hence, comparing to the original DA, the hybrid algorithm-MHDA showed better performance because the MHDA provided good stability between exploration and exploitation capabilities offered by DA and PSO respectively. Then the *pbest* and *gbest* of PSO were initialized using DA-*pbest* and DA-*gbest* matrixes respectively.

The equations for position and velocity of PSO were modified as follows:

$$V_{k+1}^i = wV_k^i + C_1 r_1 (DA - pbest_k^i - X_k^i) + C_2 r_2 (DA - gbest_k^g - X_k^i) \quad (12)$$

$$X_{k+1}^i = X_k^i + V_{k+1}^i \quad (13)$$

Where $DA - pbest_k^i$ and $DA - gbest_k^g$ are the *pbest* and *gbest* in PSO respectively. $k$ is the size of the swarm.

In [29], DA was combined with an extreme learning machine (ELM) to overcome the problems in gradient-based algorithms. In this technique to optimally select the biases of the hidden layer, DA was used. Using DA improved the ELM's overall performance. The convergence of DA-ELM was expected in a small number of iterations. Moreover, the over-fitting problems in traditional ELM overcame using the DA-ELM model. The results showed that in general DA-ELM could outperform both GA-ELM and PSO-ELM. It was also examined that DA has a good ability in searching the feature space adaptively and showed its capability in

```
Initializing the set of parameters:
Maximum iteration (Max-iter), maximum number of search agents (N
max) number of search agents (N), number of dimensions (d), upper
bound and lower bound of variables Initialize the dragonflies populations
(X) – Initialize the step vectors (ΔX)
while maximum iterations not done
    For each dragonfly Calculate fitness value
        if Fitness Value < DA-pbest
            in this iteration move the current value to DA-pbest matrix
        end if
        if fitness value < DA-gbest
            set current value as DA-gbest
        end if
    end
    For each dragonfly
        Update the food source and enemy
        Update w, s, a, c, f, and e
        Calculate S, A, C, F, and E using Eqs. (2) To (6)
        Update neighboring radius
        if a dragonfly has at least one neighboring dragonfly
            Update velocity vector using Eq. (8) Update
            position vector using Eq. (9)
        else
            Update position vector using Eq. (10)
        end if
        Check and correct new positions based on boundaries of variables
    end
    ------------------End of DA and Start of PSO------------------
    For each particle
        Initialize particle with DA-pbest matrix Set PSO-gbest as DA-gbest
    end
    while maximum iterations or minimum error criteria is not attained
        For each particle
            Calculate fitness value
            if fitness value < PSO-pbest in history
                set current value as the new PSO-pbest
            end if
        end
        Choose the particle with the best fitness value of all the
        particles as the PSO-gbest
        For each particle
            Calculate particle velocity according Eq. (12)
            Update particle position according Eq. (13)
        end
    end while
```

VISUAL 4: PSEUDO CODE FOR MHDA [26]

avoiding local minima that may cause premature convergence. Furthermore, it was proved that the DA has the minimum root mean square error that showed the ability of DA in finding optimal feature combination in less prediction error. To some extent, the average computational time of the DA was also compatible with PSO and GA. The compared and proposed models trained using a thousand iterations.

For the optimization process, RMSE was examined as a fitness function. And the number of iterations was used as a criterion to stop the process. The ranges of the Reference [30] proposed a hybrid version of the dragonfly algorithm with support vector regression (SVR) for online voltage stability assessment. Parameter selection in SVR highly affects its performance. The important parameters for SVR include penalty parameters $C$, non-sensitivity coefficient $\varepsilon$, and the kernel parameters. The DA was used in parameter settings of SVR, which improved the performance of the technique. For training the examined model (DFO-SVR) as an input, the voltage magnitude produced from PMU buses were utilized for various operating conditions, and the least values of voltage stability index (VSI) were used as output variables. Three statistical indices were utilized for evaluating the DFO-SVR model. Those statistical indices were correlation coefficient (R), root mean square error (RMSE), and the percentage of mean square error (PRMSE).

parameters ($C$, $\gamma$, and $\varepsilon$) were [1 1000], [0.0001 0.1], and [0.1 1], respectively. The optimal solution values of the $C$, $\varepsilon$, and $\gamma$ parameters for this work are shown in Table 1.

TABLE 1: THE OPTIMAL PARAMETERS FOR THE SVR MODEL FOUND BY USING DA [30]

| SVR parameters | Optimal values of SVR parameters | |
|---|---|---|
| | IEEE 30-bus | Algerian 59-bus |
| C | 971.9378 | 985.561 |
| γ | 0.1 | 0.1 |
| ε | 0.0001 | 0.0001 |

The produced results proved that the proposed model has a good performance for prediction.

Depending on reference [31] determining an adequate cluster radius is required for generating fuzzy rules. In this work, the radius of the cluster was between 0.2 and 0.5. The predicted outputs for the systems IEEE 30-bus and Algerian 59-bus systems were compared with the actual ones using DFO-SVR and ANFIS techniques respectively. The produced results proved that the performance of prediction for both systems was better in the DFO-SVR technique comparing to the ANFIS technique.

In reference [32], a hybrid version of the binary dragonfly algorithm with enhanced particle swarm optimization algorithm for solving the feature selection problem was examined. The proposed approach called Hybrid Binary Dragonfly Enhanced Particle Swarm Optimization Algorithm (HBDESPO). The ability of DA to secure diverse solutions and the enhanced PSO ability to converge to the best global solution produced a hybrid algorithm with better performance. In the examined hybrid technique dual exploration was used and excessive exploitation was avoided. In the HBDESPO technique, the velocities of participated algorithms updated independently. In this examined system, the K-nearest neighbor (KNN) was used as a classifier for ensuring robustness of the training data and reach better feature combinations. The proposed algorithm was tested on 20 standard datasets from the UCI repository. The datasets were divided into three sets: testing, validation, and training. The value of K in the KNN was assigned to 5 based on trial and error.

The training set was used for evaluating the KNN on the validation set by using the proposed technique to advise the feature selection process. For the final evaluation of the best nominated feature the training set was used. The minimization problem for this work is shown in equation (14). The setting of the optimizer and global specific parameters are shown in Table 2.

$$fitness = \alpha ER(D) + \beta \frac{|R|}{|C|} \quad (14)$$

Where *ER(D)* is the classifier's error rate, *R* represents the selected feature's length, and *C* shows the total number of features. β and α are constants for controlling the weights of classification accuracy of the minimization feature.

TABLE 2: PARAMETER SETTINGS FOR BINARY HYBRID HBDESPO [32]

| Parameter | Value |
| --- | --- |
| No. of Iterations | 70 |
| No. of search agents | 5 |
| Dimension | No. of features in the data |
| Search domain | [0 1] |
| No. of runs | 10 |
| $w_{max}$ | 0.9 |
| $w_{min}$ | 0.4 |
| $Deltax_{max}$ | 6 |
| $c_1$ | 2 |
| $c_2$ | 2 |
| $v_{max}$ | 6 |
| β in fitness function | 0.01 |
| α in fitness function | 0.99 |

The proposed technique compared to the results of binary DA from [9] and the enhanced PSO from [33]. From this research work, it was concluded that the examined algorithm could provide high classification accuracy while keeping the ratio of feature selection to the minimum. Moreover, small fitness values were reached and across various runs, the algorithm kept its stability. It was also concluded that the values of the standard deviation observed the robustness of the algorithm as it repeatedly could converge to a similar solution.

In DA, having an excessive number of social interactions may reduce the accuracy of the solution, fall easily into local optima, and cause imbalance between exploitation and exploration. To control these deficiencies, in reference [34] DA was merged with an improved version of Nelder-Mead algorithm (INMDA). The reason for this hybridization was to make the capability of local explorative stronger and prevent falling into local optima. INMDA consists of two stages. First, to search the solution space, DA was utilized and gave the required diversity to the individuals to find the global optimum solution. Second, the improved version of the Nelder-Mead (INM) simplex method was utilized to find the best and worst points and calculate the population centroid. One of the main features of INM is that the population's centroid was used to update the position. This improves the possibilities of jumping out of local optima. The efficiency of the proposed technique was tested using 19 unconstraint and 13 large-scale benchmark functions with dimensions of 30, 500 and 1000 respectively. Table 3 shows the parameter settings for INMDA. In Table 3, *t* represents the current iteration, and *T* is the number of iterations.

TABLE 3: PARAMETER SETTINGS FOR INMDA [34]

| | |
| --- | --- |
| α = 2.rand.(0.1-0.2t/T) | β = 0.9-0.5t/T |
| γ = 0.1.(1-0.2t/T) | a = 2.rand.(0.1-0.2t/T) |
| c = 0.2.rand.(1-2t/T) | e = 0.1.(1-2t/T) |
| f = 2.rand | w = 0.9-0.5t/T |
| δ = 0.68 | λa = 0.9-0.5t/T |

For single-objective functions, the proposed technique was compared to other algorithms, such as Memory-based Hybrid Dragonfly Algorithm (MHDA), DA, PSO, and recently SI-based optimization algorithms, such as ALO and WOA. For each optimization algorithm, 30 independent runs were used. The number of agents and maximum number of iterations were 30 and 1000, respectively, following the literature [26]. Fried man's test functions and Wilcoxon rank sum were used to statistically test the significance of the experimental results for the 19 unconstrained benchmark functions. The results showed the great performance of the proposed work for solving high-dimensional problems comparing to the other algorithms, such as DA and MHDA while they cannot be used to solve high-dimensional problems because they easily encounter "dimensional curse". The work concluded that the INMDA owns a superior performance comparing to the other algorithms. The enhanced exploitation and exploration capabilities from using reverse learning techniques generated this great performance.

Furthermore, for enhancing the performance of optimization by DA, reference [35] examined an improved version of DA. The proposed DA is based on exponential function adaptive steps and elite opposition-based learning strategy. The elite individual was presented to construct the opposite solutions using elite opposition-based learning. The scope of the search area expanded by using the mentioned technique and it was useful for improving the capability of the global exploration of DA. Furthermore, to replace the original stochastic step, an adaptive step with exponential step was designed. The improved work named as dragonfly based on elite opposition-based learning and exponential function adaptive steps (EOEDA). The reason behind this modification was that DA sometimes has problems to solve complex optimization problems and it easily falls into local optimum, and the speed of convergence was low. The results proved that the EOEDA had better convergence accuracy and the speed of convergence was faster.

In reference [36], different features were selected using a new chaotic dragonfly algorithm (CDA). In CDA, searching iterations of DA were merged into the chaotic maps. To modify the main parameters for movement in DA ten chaotic maps were utilized to improve the convergence rate and efficiency of the DA. The proposed method was used to select features in the extracted dataset from Drug bank database, which has 6712 drugs. In this work, 553 bio-transformed drugs were used. The proposed method was utilized to assess the toxicity of hepatic drugs. The proposed model, in general, consisted of three phases: data pre-processing, feature selection, and the classification phase. In phase two, CDA was utilized to pick the features. The k-NN classifier was used to measure the goodness of the selected features. Table 4 presents the initial parameter settings for CDA.

TABLE 4: PARAMETER SETTINGS FOR CDA, D STANDS FOR DIMENSION [36].

| Parameter | Value |
|---|---|
| $\beta$ | 1.5 |
| $D$ | 31 |
| M | 50 |
| Lower bound | 1 |
| Upper bound | 31 |
| Maximum iteration | 50 |

Furthermore, the examined technique was evaluated using three different experiments. The results proved that using chaotic maps with the dragonfly algorithm produced better results. Another experiment was conducted by comparing the performance of CDA with Gauss's chaotic map with seven methods for optimization, namely, PSO, ABC, GWO, CSA, SCA, SSA, and CSO. Table 5 shows the parameters for the participated algorithms. The results proved that CDA provided better score in most of the cases. However, CDA in most cases provided minimum stability. On the other hand, examining the *p-value* proved that statistically the produced results were important, which shows the superiority of CDA comparing to the aforementioned algorithms. To examine the selected features SVM classifier with various kernel methods was used. The results showed superiority of the CDA comparing to the other techniques. Moreover, in terms of computational time, it was noted that the computational time reduced while using feature selection algorithm and that the CDA minimized the time remarkably comparing to the original DA.

TABLE 5: PARAMETER SETTINGS FOR PSO, ABC, CSO, GWO, CSA, AND SCA OPTIMIZATION ALGORITHMS [36]

| Algorithm | Parameters | Value |
|---|---|---|
| PSO | An inertia weight | 1 |
|  | An inertia weight damping ratio | 0.9 |
|  | Personal learning coefficient | 1.5 |
|  | Global learning coefficient | 2.0 |
| ABC | number of colony size | 10 |
|  | number of food source | 5 |
|  | number of limit trials | 5 |
| CSO | number of chicken updated | 10 |
|  | The percent of roosters population size | 0.15 |
|  | The percent of hens population size | 0.7 |
|  | The percent of mother hens population size | 0.05 |
| GWO | a | 2 |
| SCA | b | 2 |
| CSA | Awareness probability | 0.1 |
|  | Flight length | 0.01 |

Reference [37] proposed a method to control frequency in an islanded AC micro-grid (MG). MG is formed by integrating various sources, such as wind power generation, renewable sources of energy, and solar energy generation. In this work ABC was merged with DA. The idea behind hybrid ABC/DA (HAD) was merge exploration and exploitation ability of the DA with the exploration ability of ABC. The proposed technique consists of three components: the dynamic and static swarming behaviour in the DA and the two phases of global search in ABC. The global search was accomplished by the first component (DA phase), and local search was accomplished by the second component (onlooker phase), and the third one accomplished global search (modified scout bee phase).

All the parameters from the original DA and ABC were adapted to the merged technique with one more parameter, which is *Prob*. The added parameter was used for balancing the application of dragonfly bee, and the onlooker bee phases, and balancing between exploitation and exploration. The *prob* parameter was set to 0.1 in the carried experiments in the work, which was based on another confirmed experiment. HAD considered *D* dimensional solutions and population size of *N*. Observing the worst case, the time complexity of the iterative process of the hybridized technique was analysed as follows: In the first phase, the main operation for creating an initial population, and the complexity time was $O(ND)$. In the second phase, the stopping criteria were judged, and the time complexity was $O(1)$. In the third phase, the value of *rand* parameter was judged. If *rand* smaller than *prob*, then perform dragonfly bee phase, else perform onlooker bee phase, then perform a modified scout bee phase, the time complexity was $O(N)$. In the fourth phase, the solution was updated, and the time complexity was $O(N)$. In the fifth phase, continue with the iterations and go back to the second step. Hence, the time complexity of the examined technique was $O(ND)$.

50 iterations were used to calculate the performance of the hybrid technique.

In terms of convergence speed, the results proved that compared to the original DA and ABC, the proposed technique provided better performance and in some cases the results were comparative. In another contribution, the examined technique was used to train multi-layer perceptron (MLP) neural network. The results showed that in terms of convergence speed, achieving the best optimal value, accuracy, and avoiding local minima, the proposed hybrid technique was a better trainer for MLPs in comparison to the original version of the participated algorithms.9

## VI. Applications of DA

Due to the power of DA, enormous research applications in applied sciences have been conducted. For example, machine learning, image processing, wireless, and network applications, and some other areas. In this section, we present applications of dragonfly algorithm in the aforementioned areas. The purpose of using the DA in the applications in various areas and the results are shown in Table 6.

TABLE 6: THE PURPOSES OF USING THE DA IN VARIOUS APPLICATIONS AND ITS RESULTS.

| Reference | Purpose | Result |
|---|---|---|
| [38] | BDA helped in searching for the optimal parameter sets (kernel parameter and penalty factor) for KELM and the optimal feature subset among the feature candidates simultaneously. | BDA showed its superiority as a searching technique to find the set of optimal parameters and the optimal feature subset. |
| [41] | Multilevel segmentation of colour fundus images. | Using the DA as an optimization algorithm produced better results for segmenting colour images. |
| [42] | In a watermarking technique for the medical images, DA was utilized to select the effective pixels. | The correlation coefficient values using the DA were greater than the other techniques such as PSO, GA, and random selection. |
| [43] | Exploring the pixels of images and discovering which pixel contains significant information about the object. (DA was used as a detection model) | The DA could work as an efficient and fast object extraction from images. |
| [44] | DA was used as a parameter optimizer of SVM. Furthermore, the effect of the number of solutions and generations on the accuracy of the produced result and computation time was investigated. | It was shown that the classification error rate for the proposed work was lower than that in PSO+SVM and GA+SVM. The reason for this was that the DA parameters could be altered iteratively. Furthermore, it was shown that increasing either the number of solutions or generations decreased the rate of misclassification and rose the computational time. |
| [48] | New updating mechanism and elitism were added to the binary dragonfly algorithm. The improved technique was then used to classify different signal types of infant cry. It was used to overcome the dimensionality problem and select the most salient features. | It was noted that the improved technique reduced the percentage of error rate comparing to the original binary dragonfly algorithm. |
| [49] | The DA based artificial neural network technique was utilized for predicting the primary fuel demand in India. | The proposed model using the DA was provided with more accurate results comparing to the existing regression models. |
| [50] | Binary-BDA, multi-BDA, and ensemble learning based BDA were used for wavelength selection. | Using binary-BDA causes instability. However, stability boosted by using the multi-BDA and the ensemble learning based BDA. In addition, the computational complexity of ensemble learning based BDA was lower than the multi-BDA. |
| [51] | Instead of gradient-based techniques, DA was used for designing filters of IIR. | Using the DA prevented trapping into local optima and coefficients close to the actual value were evaluated, and the minimum mean square value was found. In addition, the superiority of the DA was proved to compare to the PSO, CSO, and BA for the aforementioned problem. |
| [52] | Dragonfly based clustering algorithm was used to focus on the scalability of internet of vehicles. | The proposed technique was compared to a comprehensive learning PSO and ant colony optimization algorithm. The results proved that in a high density and medium density the examined technique showed better and average performance respectively. However, in a low density the proposed technique performance was bad while the comprehensive learning PSO performed well. |
| [53] | Dragonfly algorithm utilized to predict the location of randomly deployed nodes in a designated area. Al, so it was used to localizing different noise percentages of distance measurement (Pn). | For range-based localization with varying Pn, dragonflies could produce fewer errors comparing to PSO. Furthermore, increasing Pn caused an increase in the distances between real and approximated nodes by DA and PSO. |
| [54] | DA was used to enlarge the life time of the RFID network. | The cluster breakage was reduced through choosing the cluster heads that had similar mobility but high leftover energy. This reduction reduced energy consuming. Hence comparing to the existing techniques the efficiency was improved. |
| [55] | DA with two selection probabilities were used as new loud balancing technique, called (FDLA). The new technique was then used to keep the stability of processing multiple tasks in the cloud environment. | The proposed technique provided the minimum load with allocating less number of tasks. |
| [58] | DA was utilized to examine the optimal sizing and location of distributed generation in radial distribution systems to reduce the power loss in the network | Comparing to the DA and WOA, MFO performed better and converged earlier. |
| [60] | In the court case assignment problem, the ability of judicial system highly depends on time and the efficiency of operation the court case. The DA was used to find the optimal solution of the assignment problem. | The DA could show superior results comparing to the FA. |
| [61] | DA was used to optimize the optimum sitting of the capacitor in different radial distribution systems (RDSs). The main aim of this study was to minimize power loss and total cost with voltage profile enhancement. | The results proved that DA-based optimization provided comparative results with GWO- and MFO-based optimization methods in terms of small number of iterations and convergence time. However it provided superior results compared to the PSO-based technique. |

### A. Image Processing

In [38], two key crucial factors for traditional ship classification; classifier design and feature selection were joined together and a novel ship classification model called BDA-KELM classification for high-resolution synthetic aperture radar (SAR) images was proposed. Dragonfly algorithm in a binary search space in this work was used as a searching technique. For the fitness function, the accuracy of the prediction of the subsequent classifier was used. Wrapper-based methods needed too much computational time, in order to overcome this drawback, the kernel extreme learning machine was operated as the elementary classifier, and the DA helped in searching for the optimal parameter sets (kernel parameter and penalty factor) for KELM and the optimal feature subset among the feature candidates simultaneously. Integrating both selecting features and classifier design on the base of DA made BDA-KELM simple. The experiment was conducted based on Terra SAR-X SAR imagery. For training, 60% of the samples were used and the rest was used as a testing dataset. The same datasets were used in ship classification using the most popular models for classification (KNN, Bayes, Back Propagation neural network (BP neural network), and Support Vector Machine (SVM). A number of experiments were conducted using different classification models. For each model, the evaluation metrics were computed to prove the superiority of BDA-KELM. Each experiment was run 10 times. To evaluate the proposed model several evaluation metrics were used. They include recall, precision, and F1-score. The classification results proved the accuracy of the proposed model, which was 97%. Thus, the performance of the examined technique for classification was better than the examined techniques in the literature.

The thresholding of histograms is a technique that is widely used for segmenting grey scale images. However, for colour images, it is not trivial because of its multilevel structure [39, 40]. For this reason, in [41] the authors tried to overcome this problem by using dragonfly optimization algorithm for doing multilevel segmentation (SADFO) of colour fundus image. The problem of multilevel segmentation was shown as an optimization problem and DA was used to solve it. The threshold values were optimized for the chromatic channels of colour funds images by exploring the solution space effectively and finding the global best solution. Kapur's entropy was used in the proposed technique. The result proved that the proposed method produces much better results compared to segmentation after changing the image to greyscale.

In the medical images, watermarking is a hot topic that gives security to the capsulated secret code to the images. Reference [42] examined a powerful watermarking technique that depends on the weight of the pixels. To determine effective pixels for watermarking, discrete wavelet transform (DWT) was utilized for extracting the low and high-frequency bands. DA used to select the effective pixels which followed the objective function based on edge level, neighbourhood strength, gradient energy, and wavelet energy (ENeGW) of the pixels. Medical retinal images were used for the experiment and patient data was used as a watermark. In terms of performance metrics, a comparative analysis was carried out in this work. The metrics used were correlation coefficient and peak signal-to-noise ratio (PSNR). In this work, it was observed that the correlation coefficient values for the examined technique comparing to the other techniques, such as random selection, PSO, and GA were greater. The proposed work obtained the correlation at a rate 0.936719 for the random noise, 0.974479 for salt and pepper noise, and 0.983073 for the rotational noise. Similarly, the PSNR results for the examined dragonfly technique under the existence of the random noise, salt and pepper noise, and rational noise were greater and they were 62.39155 dB, 62.95912 dB, and 63.02815 dB, respectively. However, for the already existing method, such as random selection, the values of PSNR with respect to the noises were 59.59593 dB, 61.48404 dB, and 59.87195 dB, the PSNR values of PSO were 60.20927 dB, 61.63731 dB, and 60.53219 dB, respectively, and the PSNR values of the genetic algorithm were 59.62668 dB, 61.46258 dB, and 59.90074 dB, respectively.

Reference [43] used dragonfly to explore the pixels in images and assess which of these pixels represent significant components of the objects. Therefore, this technique works as a detection model for finding interesting features. In this work, a fitness function was modeled to work as a detection tool to select pixels related to the shapes of objects. In each iteration, the individuals in a given population were assessed for adaptation to the environment. In the case of images, unfortunately, the essential search areas may vary in many places. Hence, a set of tree functions was examined. Results of the examined bio-inspired extraction technique showed that utilizing different components of the fitness function resulted in different selection of key-points. This was because different aspects of the image were focused on the different components. From this work, it was concluded that the proposed technique helped on efficient and fast object extraction from images.

**B. Machine Learning**

Parameters in SVM such as the kernel and penalty parameters have a great impact on accuracy and complexity of classification model. In order to decrease classification errors in [44], dragonfly optimization algorithm was used for parameter optimization of SVM. The values of kernel and penalty parameters were sent by the DA for training SVM using the training data. The bounds for searching range of penalty parameter of SVM was $C_{min} = 0.01$ and $C_{max} = 35000$, and the bounds of the searching range of σ was $σ_{min} = 0.01$ and $σ_{max} = 100$ [45]. In this work, the effects of the number of solutions on the computational time and testing error rate were investigated.

As shown in Fig. 2 (a and b), it was concluded that increasing the number of solutions decreases the rate of misclassification. However, computational time was raised. In addition to the number of dragonflies, it was also proved that the number of generations also had effects on testing error rate and computational time. Increasing the number of generation reduces the error rate to an extent after that enlarging the number of generation did not make any changes in the error rate of misclassification. Furthermore, the computational time increased with increasing the number of generations.

For testing the datasets non-parametric Wilcoxon signed rank test was used in this proposed work. However, to test the estimated error rate, 10-fold cross-validation was used.

Furthermore, it was shown that the classification error rates for the DA+SVM algorithm were lower than those in PSO+SVM algorithm [46]. The reason for this is that the DA parameters could be altered iteratively whereas PSO parameters were fixed and they had to be set first. Thus, DA automatically made the best trade-off between explorations to exploitation. Furthermore, in comparison to GA+SVM algorithm [47] in most cases, the DA-SVM produced lower classification errors.

The data for both Figs. 2a and 2b are taken from [44], and here it is shown as figures.

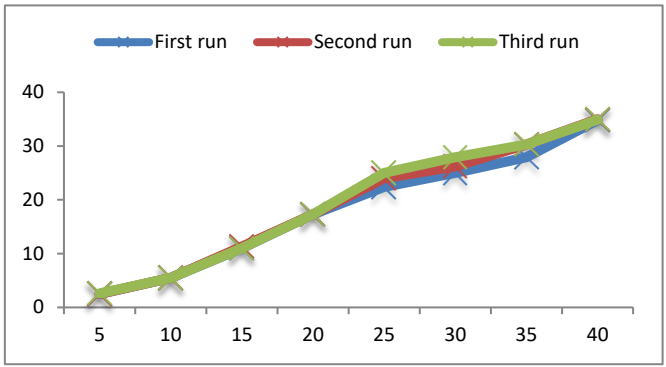

*Figure 2a: DA-SVM'S testing error rate using different number of dragonflies*

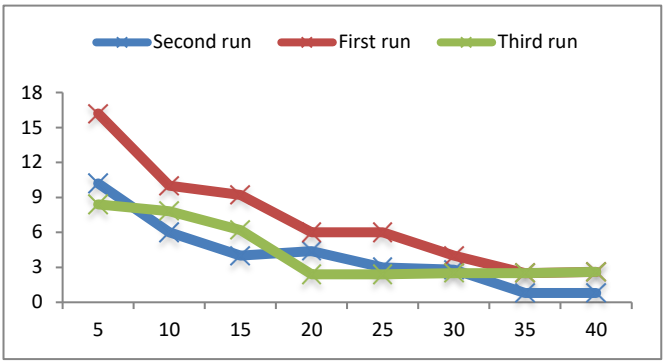

*Figure 2b: DA-SVM'S Computational time with different number of dragonflies*

Reference [48] proposed a combination technique of wavelet packet based features and the improved version of binary dragonfly optimization (IBDFO) algorithm based feature selection that was used for classifying various signal types of infant cry. Cry signals were obtained from two different databases. Each database contained a number of samples for different reasons for crying as shown in Table 7. In Mel-frequent Cepstral Coefficient (MFCCs) (16 features), Linear Predictive Coding (LPC) based cepstral (56 features); Wavelet packet transform energy and non-linear entropies (496 features) were extracted. IBDFO algorithm was used to overcome the dimensionality problems and choose the most salient features. A wrapper-based feature technique was proposed and various types of infant cry were classified. Extreme Learning Machine kernel classifier was used, all and highly informative features were utilized. New updating mechanism and elitism were added to the basic binary dragonfly optimization algorithm (BDFO) to enhance its performance when optimizing the crying features. It was noted that by using a two-class experiment, the percentage rate of recognition accuracy of IBDFO was improved very well comparing to BDFO.

It was discovered that the results achieved for IBDFO using seven class experiments were better than those achieved using other techniques (improved binary dragonfly optimization algorithm (IBDFO), GA, PSO). The results indicated that the combination of feature selection and extraction technique provided better classification accuracy.

TABLE 7: CRYING SAMPLES IN THE TWO UTILIZED DATABASES [48]

| Databases | Samples (types of the crying signal) |
|---|---|
| First Database | 507 normal crying samples (N) |
| | 340 asphyxia crying samples (A) |
| | 879 deaf crying samples (D) |
| | 350 hungry crying samples (H) |
| | 192 pain crying samples (P) |
| Second Database | 513 jaundice crying samples (J) |
| | 531 premature crying samples (Prem) |
| | 45 normal crying samples (N) |

In reference [49], DA based artificial neural network (ANN) model was utilized to estimate India's primary fuel demand. Two multi-layer feedforward networks were used. Each of the networks processed input, output, and hidden layers and each network trained with DA. Along with the networks socio-economic indicators are involved, for example, population and per capita gross domestic product (GDP). The connection weights of ANN models were optimized via searching the problem space effectively to find the global best solution. The model proposed in this work required as input, the forecast year, then the primary fuel demands are predicted. The forecast up to the year 2025 was compared with the regression model, and the forecast's accuracy was calculated using the mean absolute percent error (MAPE). This work showed that the proposed model was more accurate than the existing regression model.

Wavelength selection is a notable issue of pre-processing in near-infrared (NIR) in spectroscopy analysis and modeling. Reference [50] examined a new technique for wavelength selection based on a binary dragonfly algorithm, which consisted of three typical frameworks: multi-BDA, single-BDA, and ensemble learning based BDA settings. It was discovered that, for the aforementioned problem, using binary-BDA could cause instability. However, both ensemble learning based BDA and multi-BDA techniques could boost stability. Here, the key technical skill was to reduce the randomization inherent in the BDA. In addition, the computational complexity of multi-BDA was higher than that of the ensemble learning based BDA, which mainly has an effect on the computation of the fitness function. For performance validation of the above-mentioned techniques, the public gasoline NIR spectroscopy dataset was utilized. The aim was to observe the most representative wavelengths for predicting the content of octane. The values of the parameters of the BDA are listed in Table 8. The results proved that likewise the traditional swarm optimization techniques, the BDA could be used to deal with wavelength selection problem.

TABLE 8: PARAMETER VALUES OF THE BDA [50]

| Parameters | Values |
| --- | --- |
| Maximum No. of iterations | 50 |
| N. of dragonflies | 10 |
| No. of wavelengths | 401 |
| Separation, alignment, cohesion, food, and enemy factor | Adaptive tuning |
| No. of principal components | 2 |
| No. of folds of cross-validation | 5 |

The difference between multi-BDA and single-BDA is that a voting strategy was used to aggregate the results of the wavelength selection of the multiple-time search. From this experiment, it was found that by adjusting the vote's percentage value (VP), the number of selected wavelengths could be controlled. It was obvious that having a small number of selected wavelengths would cause a reduction in the performance of quantitative analysis model.

Moreover, in the selection of wavelength using the ensemble learning method and the BDA technique before the BDA, a serious of bootstrap sampling generators were added, which was the main difference between this technique and multi-BDA. In this work, it was shown that although the size of the sample reduced using bootstrap sampling, the performance of the quantitative analysis models with the features selected was adjacent to that of the multi-BDA method. Thus, using this technique helped in reducing the computational complexity.

Designing filters in the field of infinite impulse response (IIR) depends mainly on the conventional selection of parameters filtered among a huge possible combination. The system identification problem requires exploiting the adaptive IIR filter coefficients by using a new algorithm until it is equivalent to the examined unidentified system and adaptive filter. The design of filter for problem depends on discovering the optimal set of parameters for unrevealed model so that its close counterpart with the parameters of the filtered benchmark. In reference [51] DA was used to design the IIR filters instead of using gradient-based methods such as least mean square (LMS). Utilizing the DA prevented locating in the local optima and could evaluate the coefficients close to the actual value and the minimum mean square value was found. Furthermore, the results proved the superiority of DA against PSO, CSO, and BA to solve the aforementioned problem.

Internet of vehicles (IoV) is utilized to communicate vehicles together. As vehicular nodes are usually considered in moving, hence it makes periodic changes in the topology. Major issues are caused by these changes, such as scalability, routing shortest path, and dynamic changes in topology. Clustering is one of the solutions to such problems. Reference [52] proposed a new method calling dragonfly-based clustering algorithm (CAVDO) to focus on the scalability of IoV topology. Furthermore, mobility aware dynamic transmission range algorithm (MA-DTR) was used to transmit range adaptation based on traffic density. The proposed work was compared to comprehensive learning PSO (CLPSO) and ant colony optimization. The results proved that in a number of cases CAVDO performed better. The CAVDO performed better in a high density, an average in medium density, and performed worst in low density. However, CLPSO performed well in a very low density only.

C. Wireless and Network

Reference [53] utilized a dragonfly algorithm in two scenarios: A. To predict the location of randomly deployed nodes in a designated area. B. To localize different noise percentage of distance measurement (Pn). In both scenarios, the localization was simulated using PSO and DA. In the first scenario as shown in Fig. 3, the simulation results showed that for range based localization with varying $Pn$, dragonflies could produce fewer errors. In the second scenario, on the other hand, different numbers of unknown nodes were used for the localization; the simulation result proved that the distances between real and approximated nodes by DA and PSO were increased with the increase of Pn, see Fig. 4.

In Radio Frequency Identification (RFID) Network in order to make an improvement in energy efficiency and maximize it, the network's life should be maximized too by minimizing the use of energy RFID readers and balance the use of energy by every reader in the network. To enlarge the RFID network lifetime, DA was used in reference [54] to develop centralized, and protocoled based energy efficient cluster. A high-energy node was used as a cluster head; this devoted less amount of energy while aggregated data was transmitting to the base station. Required residual energy to receive data from the whole readers was defined as a threshold value. The readers with higher leftover energy comparing to the threshold value became the cluster head.

The data for both Figs. 3 and 4 are taken from [53], and here it is shown as figures.

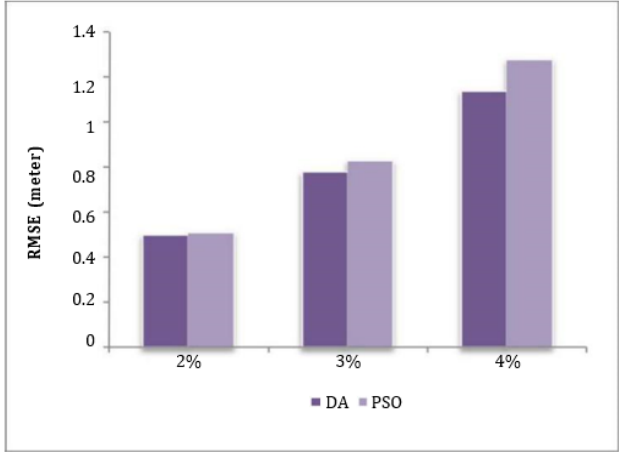

*Figure 3: Comparing RMSE between DA and PSO with varying Pn*

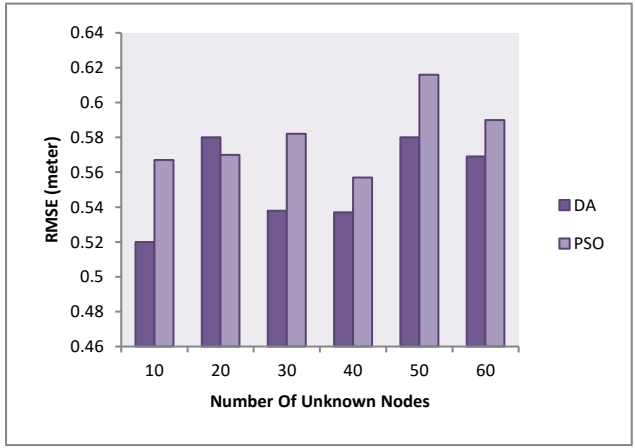

*Figure 4: Comparing RMSE between DA and PSO with different number of unknown nodes*

An optimal cluster head among all the cluster heads was chosen using dragonfly clustering. The proposed work decreased the cluster breakage by choosing the cluster head that had similar mobility but high leftover energy. Avoiding the cluster breakage decreased consuming of energy. In addition, redundancy in data was avoided in the cluster head by aggregating the data. Hence comparing to the existing methods, in the RFID network the efficiency was increased. The algorithm steps for selecting a cluster head and formation are described below:

Step 1: initializing the tags and readers in the network $R_i$

Step 2: assigning potential scores (energy and mobility) to each tag nodes and readers.

Step 3: finding the threshold value by sending the Reader's energy level to the Base station.

Step 4: if ($R_i$ > value of threshold) go to step 5 else go to step 6

Step 5: The readers are updated as Eligible Cluster Head.

Step 6: the readers are updated as remaining readers in the network.

Step 7: Separation, Alignment, and Cohesion are calculated for the eligible head in the network using equations 8, 9, and 10 respectively.

Step 8: add the values found in step 7.

Step 9: select the value from step 8 as "Optimal head" if it is high.

Step 10: else the value "become ordinary readers" in the network.

Step 11: Cluster formation ends.

In the cloud environment, keeping the stability of processing multiple tasks is a difficult issue. Therefore, a load balancing method is required to allocate the task to the virtual machines (VMs) without influencing the system's performance. Reference [55] provided a method for load balancing, named as a fractional dragonfly load balancing algorithm (FDLA). In the proposed work two selection probabilities and fractional DA were examined. FDLA was implemented by integrating fractional calculus (FC) into the process of position updating in DA. Equation (15) is the position updating equation for the proposed work. The examined model used certain parameters of physical machines (PMs) and VMs for selecting the function to be reallocated in the VMs for load balancing. Probabilities were used to select the tasks. Task selection probability (TSP) and VM selection probability (VSP) were used. The objective function for the examined technique was based on three objectives, for example, the load of VMs, task migration cost, and the capacity of VMs. The objective was to maximize the solution's fitness. The values of parameters utilized in the proposed work are shown in Table 9.

TABLE 9: VALUES OF PARAMETERS FOR FDLA [55]

| Parameters | Values |
|---|---|
| Separation weight | 0.5 |
| Alignment weight | 0.5 |
| Cohesion weight | 0.5 |
| Food factor | 0.5 |
| Enemy factor | 0.5 |
| Population size | 10 |

$$(t+1) = \alpha Y(t) + \frac{1}{2}\alpha Y(t-1) \\ + \frac{1}{6}(1-\alpha)Y(t-2) \\ + \frac{1}{24}\alpha(1-\alpha)(2-\alpha)Y(t-3) + \Delta Y(t+1) \quad (15)$$

Where, *Y(t - 1), Y(t - 2)* and *Y(t - 3)* represent the individual positions at iterations *(t - 1), (t - 2)* and *(t - 3)*, respectively, and $\Delta Y(t + 1)$ represents the step vector.

Three different techniques were compared with the examined technique in this paper. The other techniques are PSO [56], Honey Bee Behaviour inspired Load Balancing (HBB-LB) [57], and DA (DA was applied instead of the FDLA for balancing the load).

It was observed that the examined work provided a minimum load with allocating 14 tasks. Thus, the proposed FDLA obtained maximum performance than the other techniques. And the number of tasks reduced from 27 to 14 using FDLA comparing to the PSO and HBB-LB.

## VII. The Comparison Between DA and Other Algorithms

In reference [58], whale optimization algorithm (WOA), Moth-Flame optimization (MFO), and DA were compared. In the mentioned reference, these algorithms were implemented to examine the optimal sizing and location of distributed generation in radial distribution systems to reduce the power loss in the network. For this, multiple-DG units were allocated simultaneously and analyzed by considering two load power factors, i.e., unity and optimal. Bus systems 69 and 119 were used to test the algorithms. Four different cases were used to perform simulation as shown in Table 10.

TABLE 10: CASE STUDIES [58]

| Case # | The operation mode of DG | System |
|---|---|---|
| Case 1 | DG operating at a unity power factor | IEEE 69-bus radial distribution system |
| Case 2 | DG operating at an optimal power factor | |
| Case 3 | DG operating at a unity power factor | IEEE 119-bus radial distribution system |
| Case 4 | DG operating at an optimal power factor | |

The performance for the aforementioned optimization algorithms for cases 1, 2, 3, and 4 proved that the MFO algorithm showed its superiority comparing to WOA and DA. It performed better and converged earlier for the mentioned objective functions.

Reference [59] used DA, MFO, and WOA to optimize a nonlinear and stochastic optimization problem. The addressed problem in this work was finding the optimum allocation of capacity pricing and capacity between two individual markets for electricity. The markets were having non-identical designs and they were interconnected. One of the markets had an energy-only market, while the second one had a capacity-plus-energy market. The objective function for this problem was optimally allocating capacity by generation companies (GenCo) in a way that this allocation could be able to increase general revenue. Likewise, the independent system operator (ISO) acquires energy and capacity, thus, it could reduce the cost of the purchase. In this paper, it was discovered that the maximum value of GenCos's revenue, the capacity price, and the smallest value of ISO's purchase cost were increased with the cost of recall, probability of recall, and the load forecasting error. Furthermore, it was concluded that various algorithms required various numbers of iterations to converge. It is worth mentioning that none of the algorithms were proved its superiority with respect to the examined problem.

Reference [60] focused on court case assignment that has a great impact on improving the judicial system's efficiency. The ability of judicial system highly depends on time and the efficiency of operation of the court case. In this work, a mixed integer linear programming (MILP) was used to examine the case assignment issue in the justice court. The assignment problem objective included *N* cases that should be assigned to *M* teams and each team had the ability to do all the cases. Nevertheless, due to case specification, personal capability, and working on other cases at the time, the teams needed to spent different time to maximize or minimize the assignment problem's objective. To find the assignment problem's optimal solution DA and FA were used. Two problems were examined for uniform distribution the problems were shown as, for example, problem one (P1): effectiveness rate ($\mu_i$) = (1,90), Lower bound ($L_l$) = (1, 30), Upper bound ($U_i$) = (1, 90), and P2: effectiveness rate ($\mu_i$) = (1,90), Lower bound ($L_l$) = (1, 60), Upper bound ($U_i$) = (1, 90). The produced results proved that DA required less CPU time to find the optimal solution and an average of percent deviation for maximizing effectiveness comparing to FA.

The results showed that for 50 cases and 3 justice teams for experimental parameters: P1 (50:3,4,5) and P2 (50: 3,4,5) the results of DA were superior compared to those of FA.

In reference [61], DA, moth flame (MFA), and GWO techniques were examined to optimize the capacitor's optimum sitting in different radial distribution systems (RDSs). The factor of loss sensitivity was considered to determine the candidate buses. To validate the efficiency and effectiveness of the examined optimization techniques 33-, 69-, and 118-bus RDSs were considered. The main aim of this study was to minimize power loss and total cost with voltage profile enhancement. The results of the aforementioned optimization techniques were later compared with PSO to prove the superiority of the techniques. To ensure the equality in comparing the results of the techniques, the same initial population was selected for MFO, DA, GWO, and PSO for the 33-bus distribution system. The results proved that DA-, GWO-, and MFO-based optimization methods were much superior comparing to PSO-based technique in terms of a small number of iterations and convergence time for the examined study.

TABLE 11: METRICS USED TO EVALUATE THE PERFORMANCE OF ALGORITHMS [61].

| Metric |
| --- |
| Relative Error (RE) |
| Mean Absolute Error (MAE) |
| Root-mean-square error (RMSE) |
| Standard Deviation (Std) |
| Efficiency |

Furthermore, MFA-, DA-, and MFA-based optimization showed a higher convergence rate for 69-bus distribution system case. However, PSO was able to determine the optimal sizing and sitting of the capacitors for the 33-bus system, but it could not find the optimal solution for 69-bus system accurately. The metrics used to evaluate the performance of algorithms are shown in Table 11.

Additionally, the three algorithms DA, GWO, and MFA were evaluated using statistical tests. The parameter settings were implemented as the original references. Moreover, 35 iterations used, the population size was 20, and each algorithm run 30 times for each case. From the evaluation results for the mentioned algorithms it was observed that DA, GWO, and MFA had a tolerable root-mean-square error (RMSE). However, with respect to the other two techniques, GWO proved that it has the best values. Additionally, the stability of DA, GWO, and MFA was proved by the values of standard deviation (STD).

## VIII. Advantages and Disadvantages of DA

DA is one of the most recently developed algorithms in the area. As shown in the literature, it has been used to optimize various problems in different areas. One of the reasons that this algorithm has been able to contribute in different application is that it is very simple and easy to implement. As shown it suits applications in different areas. Furthermore, selecting the predators from the archive, the worst (most populated) hyper-sphere prevents the artificial dragonflies from searching around non-promising areas. Moreover, having few parameters for tuning is another advantage of DA. Furthermore, the convergence time of the algorithm is reasonable. Over other optimization algorithms it is more speed and it easily can be merged with other algorithms.

On the other hand, it does not have an internal memory that can lead to premature convergence to the local optimum. This disadvantage was overcome in reference [26] by proposing a novel Memory based Hybrid Dragonfly Algorithm (MHDA). Furthermore, DA is easily stuck into local optima, because it has high exploitation rate. Levy flight mechanism was utilized to model the random flying behavior of dragonflies in nature. Disadvantages of Levy flight is overflowing of the search area and interruption of random flights due to its big searching steps.

## IX. Results and Evaluations

In the original research work, three groups of classical benchmark functions were used to test the performance of the algorithm. The groups are unimodal (F1-F7), multi-modal (F8-F13) and composite test functions (F14-F23). Furthermore, the Wilcoxon ranksum test functions were used to show the significance of the results statistically as shown in Table 13. The results of F1-F19 in Tables 12 and 13 are taken from the original research work. However, the authors of this research work tested both PSO and DA for the results of F20-F23 in both tables. However, the DE and FA were tested on all the benchmark functions (F1-F23) by the authors.

As shown in Table 12, for unimodal test functions, in general, the results of the FA and DE outperformed DA and PSO. This proved that the FA and DE have superior exploitation and greater convergence speed compared to the DA and PSO. However, in comparison to the PSO, DA showed superior results. Furthermore, results of the aforementioned references in this reseach work proved the high convergence speed of the DA. As shown in reference [60], the DA and FA were used to optimize the same problem. The results of DA were superior than that of FA.

In addition, the results of the multimodal test functions proved the high exploration of the DA which helps in searching the search space. However, the results of the DE, in general, were better in this group of the test functions.

For the composite test functions the FA showed superior results comparing to the other algorithms. the DA has the third place among the four. Better than the PSO and worse than the other two. This means that the balance of exploration and exploitation of the FA algorithm is superior. The reason for this is that the exploration of the DA is higher than the exploitation rate.

For statistical test functions, the results of the DA and PSO were used. As shown in Table 13, the $p$ values of the unimodal test functions are less than 0.05, which means that the results were statistically significant. For most of the multi modal test functions, as shown in the table, the results statistically significant and less than 0.05. Moreover, the statistical results of the PSO and DA for most of the composite test functions were significant and less than 0.05.

TABLE 12: COMPARISON OF RESULTS OF THE CLASSICAL BENCHMARK FUNCTIONS BETWEEN DA, PSO, DE AND FA

| F | Meas. | DA | PSO | DE | FA |
|---|---|---|---|---|---|
| $F_1$ | Mean | 2.85E-18 | 4.2E-18 | **2.23e-19** | 1.72e-10 |
| | Std. | 7.16E-18 | 1.31E-18 | **1.75e-19** | 9.43e-10 |
| $F_2$ | Mean | 1.49E-05 | 0.003154 | **6.24e-12** | 6.01e-07 |
| | Std. | 3.76E-05 | 0.009811 | **2.2e-12** | 3.29e-06 |
| $F_3$ | Mean | 1.29E-06 | 0.001891 | 27.882386 | **1.58e-10** |
| | Std. | 2.1E-06 | 0.003311 | 12.845742 | **8.66e-10** |
| $F_4$ | Mean | **0.000988** | 0.001748 | 0.002883 | 5.913-03 |
| | Std. | **0.002776** | 0.002515 | 0.000577 | 0.029813 |
| $F_5$ | Mean | 7.600558 | 63.45331 | 9.50058 | **2.383765** |
| | Std. | 6.786473 | 80.12726 | 3.204155 | **1.350716** |
| $F_6$ | Mean | 4.17E-16 | 4.36E-17 | **2.22e-19** | 1.9e-10 |
| | Std. | 1.32E-15 | 1.38E-16 | **1.47e-19** | 1.04e-09 |
| $F_7$ | Mean | 0.010293 | 0.005973 | 0.005256 | **1.57e-04** |
| | Std. | 0.004691 | 0.003583 | 0.001649 | **1.01e-04** |
| $F_8$ | Mean | **-2857.58** | -7.1E+11 | - | - |
| | Std. | 383.6466 | 1.2E+12 | 30.0487686 | 239.113661 |
| $F_9$ | Mean | 16.01883 | 10.44724 | **7.31e-11** | 7.462188 |
| | Std. | 9.479113 | 7.879807 | **1.04e-10** | 4.41686 |
| $F_{10}$ | Mean | 0.23103 | 0.280137 | **2.1e-10** | 8.47e-07 |
| | Std. | 0.487053 | 0.601817 | **9.14e-11** | 4.64e-06 |
| $F_{11}$ | Mean | 0.193354 | 0.083463 | **0.001259** | 0.053309 |
| | Std. | 0.073495 | 0.035067 | **0.002957** | 0.053615 |
| $F_{12}$ | Mean | 0.031101 | 8.57E-11 | **2.32e-20** | 1.92e-12 |
| | Std. | 0.098349 | 2.71E-10 | **3.1e-20** | 1.05e-11 |
| $F_{13}$ | Mean | 0.002197 | 0.002197 | **4.25e-20** | 8.21e-12 |
| | Std. | 0.004633 | 0.004633 | **4.52e-20** | 4.5e-11 |
| $F_{14}$ | Mean | 103.742 | 150 | 0.99800 | **0.99800** |
| | Std. | 91.24364 | 135.4006 | 0 | **1.700065e-** |
| $F_{15}$ | Mean | 193.0171 | 188.1951 | 0.000698 | **3.77e-04** |
| | Std. | 80.6332 | 157.2834 | 1.546e-04 | **1.853-04** |
| $F_{16}$ | Mean | 458.2962 | 263.0948 | **-1.031628** | -1.031628 |
| | Std. | 165.3724 | 187.1352 | **6.77e-16** | 1.06e-15 |
| $F_{17}$ | Mean | 596.6629 | 466.5429 | **0.3978873** | 3.0 |
| | Std. | 171.0631 | 180.9493 | **0** | 6.05e-15 |
| $F_{18}$ | Mean | 229.9515 | 136.1759 | 2.9999999 | **-3.862782** |
| | Std. | 184.6095 | 160.0187 | 1.27e-15 | **2.79e-15** |
| $F_{19}$ | Mean | 679.588 | 741.6341 | **-3.862782** | -3.259273 |
| | Std. | 199.4014 | 206.7296 | **2.71e-15** | 0.059789 |
| $F_{20}$ | Mean | -3.32199 | -3.27047 | -3.321797 | -9.316829 |
| | Std. | -3.38E- | 0.059923 | 0.001054 | 2.21393 |
| $F_{21}$ | Mean | -10.1532 | -7.3874 | -9.867489 | -10.147907 |
| | Std. | **6.60E-15** | 3.11458 | 0.722834 | 1.396876 |
| $F_{22}$ | Mean | -10.4029 | -8.5305 | -10.381587 | -9.398946 |
| | Std. | **1.51E-06** | 3.038572 | 0.075194 | 1.99413 |
| $F_{23}$ | Mean | -10.5364 | **-9.1328** | -10.530836 | -10.2809 |
| | Std. | **2.97E-07** | 2.640148 | 0.02909 | 1.39948 |

In the original research work, the DA was not evaluated for large scale optimization problems using the CEC-C06 benchmark functions. For most of the real-world problems time is not important as much as providing an accurate answer. In addition, in reality people run an algorithm more than one trail. Which means users try to find the most successful technique to their scenario regardless of time. The 100-digit challenge also known as "CEC-C06 benchmark test functions" examine this feature of number optimization process [62]. In this survey, the algorithm is tested using the CEC-C06 2019 test functions. The utilized CEC-C06 2019 test functions are shown in Table 15. All the CEC test functions are scalable. Furthermore, the CEC04 to CEC10 is shifted, rotated, and these function are set as minimization problems that have dimension of 10, however, the CEC01 to CEC03 is not shifted and rotated, and these functions have different dimensions.

TABLE 13: THE WILCOXON RANKSUM TEST OVERALL RUNS FOR THE CLASSICAL BENCHMARK FUNCTIONS

| F | DA | PSO |
|---|---|---|
| F1 | N/A | 0.045155 |
| F2 | N/A | 0.121225 |
| F3 | N/A | 0.003611 |
| F4 | N/A | 0.307489 |
| F5 | N/A | 0.10411 |
| F6 | 0.344704 | N/A |
| F7 | 0.021134 | N/A |
| F8 | 0.000183 | N/A |
| F9 | 0.364166 | N/A |
| F10 | N/A | 0.472676 |
| F11 | 0.001008 | N/A |
| F12 | 0.140465 | N/A |
| F13 | N/A | 0.79126 |
| F14 | N/A | 0.909654 |
| F15 | 0.025748 | 0.241322 |
| F16 | 0.01133 | N/A |
| F17 | 0.088973 | N/A |
| F18 | 0.273036 | 0.791337 |
| F19 | N/A | 0.472676 |
| F20 | 0.938062 | 0.938062 |
| F21 | N/A | N/A |
| F22 | 0.256157 | 0.256157 |
| F23 | 0.59754 | 0.59754 |

Since in the original paper the results of the DA were compared to the PSO, hence, the results of DA for the CEC-C06 test functions will be compared with PSO. 100 iterations and 30 agents were used for both algorithms. The results are presented in Table 14. As shown, the results of both algorithms are comparative in CEC03 and CEC10. however, the DA provided better results in CEC01, CEC02, and CEC06. in the rest of the CEC functions the results of the PSO were better.

The results of the CEC-C06 2019 benchmark functions proved that the DA algorithm can be used to solve large scale optimization problems.

## X. Discussion and Future Works

Unlike evolutionary algorithms and similar to other swarming techniques, the DA algorithm has few parameters for adjusting, this makes it easier to implement the algorithm. It provides a good optimization capability. As proved in the aforementioned references, DA has become a strong metaheuristic algorithm to address complex problems in most of the cases. It also provides a good convergence towards global optima.

Furthermore, the superiority and effectiveness of this algorithm were proved by the applications that utilized this technique.

TABLE 14: IEEE CEC-C06 2019 BENCHMARK TEST RESULTS

| CEC | Meas. | DA | PSO |
|---|---|---|---|
| F1 | Mean | **46835.63679** | 1.47127E+12 |
|    | Std.  | **8992.755502** | 1.32362E+12 |
| F2 | Mean | **18.31681239** | 15183.91348 |
|    | Std.  | **0.041929318** | 3729.553229 |
| F3 | Mean | 12.70240422 | 12.70240422 |
|    | Std.  | 1.50E-12 | 9.03E-15 |
| F4 | Mean | 103.3295366 | **16.80077558** |
|    | Std.  | 20.00405422 | **8.199076134** |
| F5 | Mean | 1.177303105 | **1.138264955** |
|    | Std.  | 0.057569859 | **0.089389848** |
| F6 | Mean | 5.646572343 | 9.305312443 |
|    | Std.  | **4.27E-08** | 1.69E+00 |
| F7 | Mean | 898.5188217 | **160.6863065** |
|    | Std.  | 4.023921424 | 104.2035197 |
| F8 | Mean | 6.210996106 | **5.224137165** |
|    | Std.  | 0.001657324 | 0.786760649 |
| F9 | Mean | 2.601134198 | **2.373279266** |
|    | Std.  | 0.233292964 | **0.018437068** |
| F10 | Mean | 20.0506995 | 20.28063455 |
|     | Std.  | 0.070920925 | 0.128530895 |

The DA uses low cohesion weight and high alignment for exploring the search space. For exploiting the search space, on the other hand, low alignment and high cohesion weights are used. Another technique to balance exploration and exploitation is tuning the swarming weights $s, a, c, f, e$, and $w$) adaptively during the optimization process. To switch between exploitation and exploration the radii of neighborhood enlarged proportionally to the iteration numbers can be used. For small and medium scale problems the DA usually can produce good results. For large scale problems, however, it needs more affords.

One of the difficulties that may face the users of the DA is that position updating and population centroid of the algorithm are not correlated. This may cause trapping into local optima and difficulty in finding global optima, and solutions with low accuracy. As mentioned in [18] the performance of the cuckoo algorithm was improved in a number of references by changing the levy flight mechanism. Hence, testing other strategies instead of the levy flight in the DA is highly recommended.

Moreover, the exploration and exploitation of the DA algorithm are mainly determined by alignment, separation, cohesion, and attraction toward food sources and distraction toward enemy sources. This technique improved diversity of solutions and caused exploration of the algorithm to become stronger. Nevertheless, the performance of the algorithm decreased with a lot of operators of exploration and exploitation because they cause an increase in the convergence time and trapping into local optima. As discussed in [34], for complex optimization problems, the DA easily falls into local optima and the convergence speed is low. However, for simple problems, the static swarming behaviour of the DA increases the exploration level of the algorithm and helps in avoiding local optima. Furthermore, increasing the number of iterations will result in high exploitation degree and will increase accuracy in finding approximate global optimum.

Additionally, evaluating the algorithm in the previous section proved that the DA may have problems in balancing exploration and exploitation in some cases, this was because exploration of the DA is high. In the early steps of the optimization process, the high rate of exploration is good, however, it should be decreased in the final steps of the process and the exploitation rate should be increased. Contrarily, the results of the unimodal test functions and the results of most of the reviewed research works proved the superior convergence of the algorithm for simple to medium problems.

To overcome the bottlenecks of the algorithm, it was hybridized with other algorithms. For instance, MHDA was proposed to overcome the problem of premature convergence to local optima. In spite of the fact that the DA and its hybridized versions have been able to provide good results in solving a number of complex optimization problems, some drawbacks still exist. In dragonfly algorithm, attraction towards food and distraction towards enemies provide a high capacity of exploration and exploitation during optimization technique. Nevertheless, the correlation of position updating rule of DA with the centroid of the population from the previous generation is less. Thus, this may result in a solution with low accuracy, premature convergence to local optima, and difficulties in finding the global optima. Hence, researches are encouraged to find new techniques to update the positions of dragonflies. Furthermore, another point that would help in improving the algorithm is balancing the exploration and the exploitation phases of the algorithm. This would prevent the algorithm to trap into the local optima. Furthermore, combining new search techniques with the DA and examining new transfer functions with the binary DA are highly recommended. We recommend integrating DA with other methods to dynamically tune the parameters during the optimization process. This technique would be able to provide a better balance between exploration and exploitation.

## XI. Conclusions

In this paper, one of the most recently developed algorithms was reviewed. The different variants of the algorithms including the hybridization versions with other algorithms were discussed. Furthermore, convergence, exploration, and exploitation of the algorithm were addressed. In addition, a number of optimization problems and applications that used DA were reviewed. During the research, it has been discovered that the DA in most of the cases has a good ability to converge towards the global optimum. Moreover, it has the ability to optimize different and complex problems in various areas. Additionally, the results of the benchmark functions proved that the DA has a great ability in solving simple to medium problems. However, for complex problems it may face some difficulties during the optimization. These difficulties were overcome by merging the algorithm with other algorithms. Moreover, it was also shown that the balance of exploration and exploitation of the algorithm is not good. However, in some applications it was shown that the balance of exploration and exploitation of the algorithm is reasonable. For example, for optimizing the parameters in the analyzing stress of perforated orthotropic plates, DA outperformed GA and PSO and it converged earlier since it has higher exploration and exploitation rate. Additionally, DA was successfully used to develop a new method to solve economic dispatch incorporating solar energy. The results proved that the economy could be raised and at the same time system loss could be minimized. In a binary searc space, DA was successfully used as a searching techniques. Furthermore, compared to GA and PSO, BDA showed a better searching ability and showed the ability to select features with more information. Furthermore, for doing multilevel segmentation for colour fundus image, DA showed its superiority over the other techniques that required changing the image to colour scale before doing the segmentation

# Appendix 1

TABLE 15: CEC-C06 2019 BENCHMARK FUNCTIONS [62]

| Function | Functions | Dimension | Range | $f_{min}$ |
|---|---|---|---|---|
| CE01 | STORN'S CHEBYSHEV POLYNOMIAL FITTING PROBLEM | 9 | [-8192, 8192] | 1 |
| CEC02 | INVERSE HILBERT MATRIX PROBLEM | 16 | [-16384, 16384] | 1 |
| CEC03 | LENNARD-JONES MINIMUM ENERGY CLUSTER | 18 | [-4, 4] | 1 |
| CEC04 | RASTRIGIN'S FUNCTION | 10 | [-100, 100] | 1 |
| CEC05 | GRIENWANK'S FUNCTION | 10 | [-100, 100] | 1 |
| CEC06 | WEIERSRASS FUNCTION | 10 | [-100, 100] | 1 |
| CEC07 | MODIFIED SCHWEFEL'S FUNCTION | 10 | [-100, 100] | 1 |
| CEC08 | EXPANDED SCHAFFER'S F6 FUNCTION | 10 | [-100, 100] | 1 |
| CEC09 | HAPPY CAT FUNCTION | 10 | [-100, 100] | 1 |
| CEC10 | ACKLEY FUNCTION | 10 | [-100, 100] | 1 |

## Conflicts of Interest

The authors declare that there is no conflict of interest regarding the publication of this paper.

## Funding Statement

This research did not receive any specific grant from funding agencies in the public, commercial, or not-for-profit sectors.

## References

[1] Yang, X. and He, X. (2014). Swarm Intelligence and Evolutionary Computation: Overview and Analysis. *Studies in Computational Intelligence*, [online] 585, pp.1-23. Available at: https://link.springer.com/chapter/10.1007/978-3-319-13826-8_1 [Accessed 1 Mar. 2019].

[2] Dorigo M., (1992) "Optimization, Learning and Natural Algorithms", PhD thesis [in Italian], Dipartimento di Elettronica, Politecnico di Milano, Milan, Italy.

[3] Kothari, V., Anuradha, J., Shah, S. and Mittal, P. (2012). A Survey on Particle Swarm Optimization in Feature Selection. *Communications in Computer and Information Science*, [online] pp.192-201. Available at: https://link.springer.com/chapter/10.1007/978-3-642-29216-3_22 [Accessed 2 Feb. 2018].

[4] Zhang, Y., Wang, S. and Ji, G. (2015). *A Comprehensive Survey on Particle Swarm Optimization Algorithm and Its Applications*. [online] Available at: https://www.hindawi.com/journals/mpe/2015/931256/ [Accessed 4 Feb. 2018].

[5] Kennedy, J. and Eberhart, R. (1995). Particle swarm optimization. *Proceedings of ICNN'95 - International Conference on Neural Networks*. [online] Available at: https://ieeexplore.ieee.org/document/488968 [Accessed 7 Feb. 2018].

[6] Dorigo, M. and Carro, D. (2002). Ant colony optimization: a new meta-heuristic. In: Proceedings of the 1999 Congress on Evolutionary Computation-CEC99. [online] Washinghton, DC, USA: IEEE. Available at: https://ieeexplore.ieee.org/document/782657/ [Accessed 1 May 2018].

[7] Chu, S., Tsai, P. and Pan, J. (2006). Cat Swarm Optimization. *Lecture Notes in Computer Science*, [online] pp.854-858. Available at: https://link.springer.com/chapter/10.1007/978-3-540-36668-3_94 [Accessed 6 Jan. 2018].

[8] Mirjalili, S., Mirjalili, S. and Lewis, A. (2014). Grey Wolf Optimizer. *Advances in Engineering Software*, [online] 69, pp.46-61. Available at: https://www.sciencedirect.com/science/article/pii/S0965997813001853 [Accessed 3 Jan. 2018].

[9] Mirjalili, S. (2015). Dragonfly algorithm: a new meta-heuristic optimization technique for solving single-objective, discrete, and multi-objective problems. *Neural Computing and Applications*, [online] 27(4), pp.1053-1073. Available at: https://link.springer.com/article/10.1007/s00521-015-1920-1 [Accessed 2 Jan. 2018].

[10] Storn, R. and Price, K. (1997). Differential Evolution – A Simple and Efficient Heuristic for global

Optimization over Continuous Spaces. *Journal of Global Optimization*, [online] 11(4), pp.341-359. Available at: https://link.springer.com/article/10.1023/A:1008202821328 [Accessed 2 Jan. 2018].

[11] Karaboga, D. and Basturk, B. (2007). A powerful and efficient algorithm for numerical function optimization: artificial bee colony (ABC) algorithm. *Journal of Global Optimization*, [online] 39(3), pp.459-471. Available at: https://link.springer.com/article/10.1007/s10898-007-9149-x [Accessed 6 Feb. 2018].

[12] Abdullah, J. and Ahmed, T. (2019). Fitness Dependent Optimizer: Inspired by the Bee Swarming Reproductive Process. *IEEE Access*, 7, pp.43473-43486.

[13] Shamsaldin, A., Rashid, T., Al-Rashid Agha, R., Al-Salihi, N. and Mohammadi, M. (2019). Donkey and smuggler optimization algorithm: A collaborative working approach to path finding. *Journal of Computational Design and Engineering*, [online] 6(4), pp.562-583. Available at: https://www.sciencedirect.com/science/article/pii/S2288430018303178 [Accessed 1 May 2019].

[14] Yang, X. (2009). Firefly Algorithms for Multimodal Optimization. *Stochastic Algorithms: Foundations and Applications*, [online] pp.169-178. Available at: https://link.springer.com/chapter/10.1007/978-3-642-04944-6_14 [Accessed 8 Nov. 2019].

[15] Mohammed, H., Umar, S. and Rashid, T. (2019). A Systematic and Meta-Analysis Survey of Whale Optimization Algorithm. *Computational Intelligence and Neuroscience*, 2019, pp.1-25.

[16] Dokeroglu, T., Sevinc, E., Kucukyilmaz, T. and Cosar, A. (2019). A survey on new generation metaheuristic algorithms. *Computers & Industrial Engineering*, 137, p.106040.

[17] Rajakumar, R., Dhavachelvan, P. and Vengattaraman, T. (2016). A survey on nature inspired meta-heuristic algorithms with its domain specifications. *2016 International Conference on Communication and Electronics Systems (ICCES)*. [online] Available at: https://ieeexplore.ieee.org/document/7889811 [Accessed 8 Nov. 2019].

[18] Chiroma, H., Herawan, T., Fister, I., Fister, I., Abdulkareem, S., Shuib, L., Hamza, M., Saadi, Y. and Abubakar, A. (2017). Bio-inspired computation: Recent development on the modifications of the cuckoo search algorithm. *Applied Soft Computing*, [online] 61, pp.149-173. Available at: https://www.sciencedirect.com/science/article/abs/pii/S1568494617304738 [Accessed 8 Nov. 2019].

[19] Yang, X. and Suash Deb (2009). Cuckoo Search via Lévy flights. *2009 World Congress on Nature & Biologically Inspired Computing (NaBIC)*. [online] Available at: https://ieeexplore.ieee.org/document/5393690 [Accessed 8 Nov. 2019].

[20] Russell RW, May ML, Soltesz KL, Fitzpatrick JW (1998). Massive swarm migrations of dragonflies (Odonata) in eastern North America. *The American Midland Naturalist,* 140(2), pp.325–342. Available at: https://www.jstor.org/stable/2426949?seq=1#metadata_info_tab_contents [Accessed 14 Feb. 2018].

[21] Reynolds CW. (1987). Flocks, herds and schools: a distributed behavioral model. *ACM SIGGRAPH Comput Gr,* 21(4), pp.25–34. Available at: https://dl.acm.org/citation.cfm?id=37406 [Accessed 14 Feb. 2018].

[22] Mirjalili, S. and Lewis, A. (2013). S-shaped versus V-shaped transfer functions for binary Particle Swarm Optimization. *Swarm and Evolutionary Computation*, [online] 9, pp.1-14. Available at: https://www.sciencedirect.com/science/article/abs/pii/S2210650212000648 [Accessed 2 Mar. 2019].

[23] Mirjalili, S. and Lewis, A. (2015). Novel performance metrics for robust multi-objective optimization algorithms. *Swarm and Evolutionary Computation*, [online] 21, pp.1-23. Available at: https://www.sciencedirect.com/science/article/abs/pii/S2210650214000777 [Accessed 22 Feb. 2018].

[24] Coello Coello, C. (2009). Evolutionary multi-objective optimization: some current research trends and topics that remain to be explored. *Frontiers of Computer Science in China*, [online] 3(1), pp.18-30. Available at: https://link.springer.com/article/10.1007/s11704-009-0005-7 [Accessed 22 Feb. 2018].

[25] Coello, C., Pulido, G. and Lechuga, M. (2004). Handling multiple objectives with particle swarm optimization. *IEEE Transactions on Evolutionary Computation*, [online] 8(3), pp.256-279. Available at: https://ieeexplore.ieee.org/document/1304847 [Accessed 9 Mar. 2018].

[26] K.S., S. and Murugan, S. (2017). Memory based Hybrid Dragonfly Algorithm for numerical optimization problems. *Expert Systems with Applications*, [online] 83, pp.63-78. Available at: https://www.sciencedirect.com/science/article/pii/S0957417417302762 [Accessed 9 Mar. 2018].

[27] Parouha, R. and Das, K. (2016). A memory based differential evolution algorithm for unconstrained optimization. *Applied Soft Computing*, [online] 38, pp.501-517. Available at: https://www.sciencedirect.com/science/article/pii/S1568494615006602 [Accessed 5 Mar. 2018].

[28] Ma, H., Simon, D., Fei, M., Shu, X. and Chen, Z. (2014). Hybrid biogeography-based evolutionary algorithms. *Engineering Applications of Artificial Intelligence*, [online] 30, pp.213-224. Available at: https://www.sciencedirect.com/science/article/abs/pii/S0952197614000189 [Accessed 5 Mar. 2018].

[29] Salam, M., Zawbaa, H., Emary, E., Ghany, K. and Parv, B. (2016). A hybrid dragonfly algorithm with extreme learning machine for prediction. *2016 International Symposium on INnovations in Intelligent SysTems and Applications (INISTA)*. [online] Available at: https://ieeexplore.ieee.org/document/7571839 [Accessed 11 Mar. 2018].

[30] Amroune M, Bouktir T, Musirin I (2018) Power system voltage stability assessment using a hybrid approach combining dragonfly optimization algorithm and support vector regression. Arab J Sci Eng. Available at: https://doi.org/10.1007/s13369-017-3046-5 [Accessed 24 Feb. 2019].

[31] Gopalakrishnan, K., Attoh-Okine, N. and Ceylan, H. (2009). *Intelligent and Soft Computing in Infrastructure Systems Engineering*. Berlin: Springer.

[32] A. Tawhid, M. and B. Dsouza, K. (2018). Hybrid binary dragonfly enhanced particle swarm optimization algorithm for solving feature selection problems. *Mathematical Foundations of Computing*, [online] 1(2), pp.181-200. Available at: http://aimsciences.org/article/doi/10.3934/mfc.2018009?viewType=html [Accessed 2 Mar. 2019].

[33] Mirjalili, S. and Lewis, A. (2013). S-shaped versus V-shaped transfer functions for binary Particle Swarm Optimization. *Swarm and Evolutionary Computation*, [online] 9, pp.1-14. Available at: https://www.sciencedirect.com/science/article/abs/pii/S2210650212000648 [Accessed 2 Mar. 2019].

[34] Xu, J. and Yan, F. (2018). Hybrid Nelder–Mead Algorithm and Dragonfly Algorithm for Function Optimization and the Training of a Multilayer Perceptron. *Arabian Journal for Science and Engineering*. [online] Available at: https://link.springer.com/article/10.1007%2Fs13369-018-3536-0 [Accessed 2 Mar. 2019].

[35] Song, J. and Li, S. (2017). Elite opposition learning and exponential function steps-based dragonfly algorithm for global optimization. *2017 IEEE International Conference on Information and Automation (ICIA)*. [online] Available at: https://ieeexplore.ieee.org/abstract/document/8079080 [Accessed 2 Mar. 2019].

[36] Sayed, G., Tharwat, A. and Hassanien, A. (2018). Chaotic dragonfly algorithm: an improved metaheuristic algorithm for feature selection. *Applied Intelligence*, [online] 49(1), pp.188-205. Available at: https://link.springer.com/article/10.1007%2Fs10489-018-1261-8.

[37] Khadanga, R., Padhy, S., Panda, S. and Kumar, A. (2018). Design and Analysis of Tilt Integral Derivative Controller for Frequency Control in an Islanded Microgrid: A Novel Hybrid Dragonfly and Pattern Search Algorithm Approach. *Arabian Journal for Science and Engineering*, [online] 43(6), pp.3103-3114. Available at: https://link.springer.com/article/10.1007/s13369-018-3151-0 [Accessed 12 Mar. 2019].

[38] Wu, J., Zhu, Y., Wang, Z., Song, Z., Liu, X., Wang, W., Zhang, Z., Yu, Y., Xu, Z., Zhang, T. and Zhou, J. (2017). A novel ship classification approach for high resolution SAR images based on the BDA-KELM classification model. *International Journal of Remote Sensing*, [online] 38(23), pp.6457-6476. Available at: https://www.tandfonline.com/doi/abs/10.1080/01431161.2017.1356487 [Accessed 19 Apr. 2018].

[39] Dong, G. and Xie, M. (2005). Color Clustering and Learning for Image Segmentation Based on Neural Networks. I*EEE Transactions on Neural Networks*, 16(4), pp.925-936.

[40] Sağ, T. and Çunkaş, M. (2015). Color image segmentation based on multiobjective artificial bee colony optimization. *Applied Soft Computing*, [online] 34, pp.389-401. Available at: https://www.sciencedirect.com/science/article/abs/pii/S1568494615003166 [Accessed 19 Apr. 2018]

[41] Rakoth, S. and Sasikala, J. (2017). Multilevel Segmentation of Fundus Images using Dragonfly Optimization. *International Journal of Computer Applications*, [online] 164(4), pp.28-32. Available at: https://pdfs.semanticscholar.org/5469/997649097c213061bc71ca1dc96285b3c649.pdf

[42] Hemamalini, B. and Nagarajan, V. (2018). Wavelet transform and pixel strength-based robust watermarking using dragonflyoptimization. *Multimedia Tools and Applications*. [online] Available at: https://link.springer.com/article/10.1007/s11042-018-6096-0 [Accessed 5 Mar. 2019].

[43] Połap, D. and Woźniak, M. (2017). Detection of Important Features from Images Using Heuristic Approach. *Communications in Computer and Information Science*, [online] pp.432-441. Available at: https://link.springer.com/chapter/10.1007/978-3-319-67642-5_36 [Accessed 5 Mar. 2019].

[44] Tharwat, A., Gabel, T. and Hassanien, A. (2017). Parameter Optimization of Support Vector Machine Using Dragonfly Algorithm. *Proceedings of the International Conference on Advanced Intelligent Systems and Informatics 2017*, [online] pp.309-319. Available at: https://link.springer.com/chapter/10.1007/978-3-319-64861-3_29 [Accessed 10 Apr. 2018].

[45] Tharwat, A., Hassanien, A. and Elnaghi, B. (2017). A BA-based algorithm for parameter optimization of Support Vector Machine. *Pattern Recognition Letters*, 93, pp.13-22.

[46] Lin, S., Ying, K., Chen, S. and Lee, Z. (2008). Particle swarm optimization for parameter determination and feature selection of support vector machines. *Expert Systems with Applications*, [online] 35(4), pp.1817-1824. Available at: https://www.sciencedirect.com/science/article/pii/S0957417407003752 [Accessed 2 Jul. 2018].

[47] Huang, C. and Wang, C. (2006). A GA-based feature selection and parameters optimizationfor support vector machines. *Expert Systems with Applications*, [online] 31(2), pp.231-240. Available at: https://www.sciencedirect.com/science/article/pii/S0957417405002083 [Accessed 2 Jul. 2018].

[48] Hariharan, M., Sindhu, R., Vijean, V., Yazid, H., Nadarajaw, T., Yaacob, S. and Polat, K. (2018). Improved binary dragonfly optimization algorithm and wavelet packet based non-linear features for infant cry classification. *Computer Methods and Programs in Biomedicine*, [online] 155, pp.39-51. Available at: https://www.sciencedirect.com/science/article/pii/S0169260717307666 [Accessed 3 Mar. 2018].

[49] Kumaran, J. and Sasikala, J. (2017). Dragonfly Optimization based ANN Model for Forecasting India's Primary Fuels' Demand. *International Journal of Computer Applications*, [online] 164(7), pp.18-22. Available at: https://pdfs.semanticscholar.org/e4e2/d58bda7809398c3001ea650ab54abe57f0d1.pdf [Accessed 3 Mar. 2018].

[50] Chen, Y. and Wang, Z. (2019). Wavelength Selection for NIR Spectroscopy Based on the Binary Dragonfly Algorithm. *Molecules*, [online] 24(3), p.421. Available at: https://www.mdpi.com/1420-3049/24/3/421 [Accessed 6 Mar. 2019].

[51] Singh, S., Ashok, A., Kumar, M., Garima and Rawat, T. (2018). Optimal Design of IIR Filter Using Dragonfly Algorithm. *Advances in Intelligent Systems and Computing*, [online] pp.211-223. Available at: https://link.springer.com/chapter/10.1007/978-981-13-1819-1_21 [Accessed 5 Mar. 2018].

[52] Aadil, F., Ahsan, W., Rehman, Z., Shah, P., Rho, S. and Mehmood, I. (2018). Clustering algorithm for internet of vehicles (IoV) based on dragonfly optimizer (CAVDO). *The Journal of Supercomputing*, [online] 74(9), pp.4542-4567. Available at: https://link.springer.com/article/10.1007/s11227-018-2305-x [Accessed 5 Mar. 2018].

[53] Daely, P. and Shin, S. (2016). Range based wireless node localization using Dragonfly Algorithm. *2016 Eighth International Conference on Ubiquitous and Future Networks (ICUFN)*. [online] Available at: https://ieeexplore.ieee.org/document/7536950 [Accessed 5 Mar. 2018].

[54] Hema, C., Sankar, S. and Sandhya (2016). Energy efficient cluster based protocol to extend the RFID network lifetime using dragonfly algorithm. *2016 International Conference on Communication and Signal Processing (ICCSP)*. [online] Available at: https://ieeexplore.ieee.org/document/7754194 [Accessed 5 Mar. 2018].

[55] Ashok Kumar, C., Vimala, R., Aravind Britto, K. and Sathya Devi, S. (2018). FDLA: Fractional Dragonfly based Load balancing Algorithm in cluster cloud model. *Cluster Computing*. [online] Available at: https://link.springer.com/article/10.1007/s10586-018-1977-6 [Accessed 5 Mar. 2019].

[56] Ramezani, F., Lu, J. and Hussain, F. (2013). Task-Based System Load Balancing in Cloud Computing Using Particle Swarm Optimization. *International Journal of Parallel Programming*, [online] 42(5), pp.739-754. Available at: https://link.springer.com/article/10.1007/s10766-013-0275-4 [Accessed 15 Mar. 2019].

[57] L.D., D. and Venkata Krishna, P. (2013). Honey bee behavior inspired load balancing of tasks in cloud computing environments. *Applied Soft Computing*, [online] 13(5), pp.2292-2303. Available at: https://pdfs.semanticscholar.org/e78b/31770db76a15f68936b44a734bc39e922461.pdf [Accessed 19 Mar. 2019].

[58] Saleh, A., Mohamed, A., Hemeida, A. and Ibrahim, A. (2018). Comparison of different optimization techniques for optimal allocation of multiple distribution generation. *2018 International Conference on Innovative Trends in Computer Engineering (ITCE)*. [online] Available at: https://ieeexplore.ieee.org/abstract/document/8316644 [Accessed 23 Feb. 2019].

[59] Parmar, A. and Darji, P. (2018). Comparative Analysis of Optimum Capacity Allocation and Pricing in Power Market by Different Optimization Algorithms. *Advances in Intelligent Systems and Computing*, [online] pp.311-326. Available at: https://link.springer.com/chapter/10.1007/978-981-13-1595-4_25 [Accessed 23 Feb. 2019].

[60] Wongsinlatam, W. and Buchitchon, S. (2018). The Comparison between Dragonflies Algorithm and Fireflies Algorithm for Court Case Administration: A Mixed Integer Linear Programming. *Journal of Physics: Conference Series*, [online] 1061, p.012005. Available at: https://iopscience.iop.org/article/10.1088/1742-6596/1061/1/012005 [Accessed 23 Feb. 2019].

[61] Diab, A. and Rezk, H. (2018). Optimal Sizing and Placement of Capacitors in Radial Distribution Systems Based on Grey Wolf, Dragonfly and Moth–Flame Optimization Algorithms. *Iranian Journal of Science and Technology, Transactions of Electrical*

*Engineering*, [online] 43(1), pp.77-96. Available at: https://link.springer.com/article/10.1007/s40998-018-0071-7 [Accessed 24 Feb. 2019].

[62] Digalakis, J. and Margaritis, K. On benchmarking functions for genetic algorithms. *International Journal of Computer Mathematics* 2001; [online] 77(4), pp.481-506. Available at: https://www.tandfonline.com/doi/abs/10.1080/00207160108805080 [Accessed 1 Jun. 2019].